\documentclass[11pt,a4paper]{article}
\usepackage[hyperref]{acl2018}

\aclfinalcopy 
\def\aclpaperid{23} 

\usepackage{times,latexsym,arydshln,amsmath,graphicx,xcolor,multirow,arydshln}
\usepackage{amsfonts, amssymb}

\newcommand{\todooff}{\long\gdef\todo##1{}}
\newcommand{\todoon}{\long\gdef\todo##1{{
\bf\textcolor{red} {TODO: ##1}
}}}
\todoon
\todooff

\newcounter{notecounter}
\newcommand{\enotesoff}{\long\gdef\enote##1##2{}}
\newcommand{\enoteson}{\long\gdef\enote##1##2{{
\stepcounter{notecounter}
{\large\bf
\hspace{1cm}\arabic{notecounter} $<<<$ ##1: ##2
$>>>$\hspace{1cm}}}}}
\enoteson
\enotesoff

\def\figref#1{Figure~\ref{fig:#1}}
\def\figlabel#1{\label{fig:#1}\label{p:#1}}

\def\tabref#1{Table~\ref{tab:#1}}
\def\tablabel#1{\label{tab:#1}\label{p:#1}}

\def\eqref#1{Eq.~\ref{eqn:#1}}

\usepackage{url}

\title{Evaluating Word Embeddings \\ in Multi-label Classification \\ Using Fine-grained Name Typing}

\newcommand*{\affaddr}[1]{#1} 
\newcommand*{\affmark}[1][*]{\textsuperscript{#1}}

\author{{\centering
Yadollah Yaghoobzadeh{\rm\affmark[1]} ~~  Katharina Kann{\rm\affmark[2]} ~~ 
Hinrich Sch\"{u}tze{\rm\affmark[3]}} \vspace{.15cm}
\\
\affaddr{\affmark[1]Mirosoft Research, Montreal, Canada}\\
\affaddr{\affmark[2]Center for Data Science, New York University, USA}\\
\affaddr{\affmark[3]CIS, LMU Munich, Germany} \\
\affaddr{\texttt{yayaghoo@microsoft.com}}}

\date{}

\begin{document}

\maketitle

\begin{abstract}
Embedding models typically associate
each word with a single real-valued vector,
 representing its different properties. 
Evaluation methods, therefore, need to analyze
the accuracy and completeness of these
properties in embeddings. 
This requires
 fine-grained analysis of embedding subspaces.
Multi-label classification is an
 appropriate way to do so. 
 We propose a
 new evaluation method for word embeddings based on multi-label classification given a word embedding. 
 The task we use is fine-grained
 name typing: given a large corpus, find all types
 that a name can refer to based on the name
embedding. Given the scale of entities in
 knowledge bases, we can build datasets
 for this task that are complementary to the
 current embedding evaluation datasets in:
 they are very large, contain fine-grained
 classes, and allow the direct evaluation of
 embeddings without confounding factors
 like sentence context.

\end{abstract}
\section{Introduction}
Distributed representation of words, aka word embedding, is an important element of many natural language processing applications.
The quality of word embeddings is assessed using different methods.
\newcite{baroni14predict} evaluate word embeddings on different intrinsic tests:
similarity, analogy, synonym detection, categorization and selectional preference.
Different concept categorization datasets are introduced. 
These datasets are  small ($<$500) \cite{baroni14predict,Rubinstein15embed}
and therefore measure the goodness of embeddings by the quality of their clustering.
Usually cosine  is used as the similarity metric between embeddings, ignoring subspace similarities.
Extrinsic evaluations 
are also used, cf. \newcite{LiJ15a}.
In these tasks, embeddings are used 
in context/sentence representations with composition involved. 

In this paper, we propose 
a new evaluation method.
In contrast to the prior work on \emph{intrinsic evaluation},  our method is supervised, large-scale, fine-grained, automatically built,
and evaluates embeddings in a classification
setting where different subspaces of embeddings need to be analyzed.
In contrast to the prior work on \emph{extrinsic evaluation}, we evaluate embeddings in isolation, 
without confounding factors like sentence contexts or composition functions.

\begin{figure}[t]
\centering
\includegraphics[width=0.8\columnwidth,height=105pt]{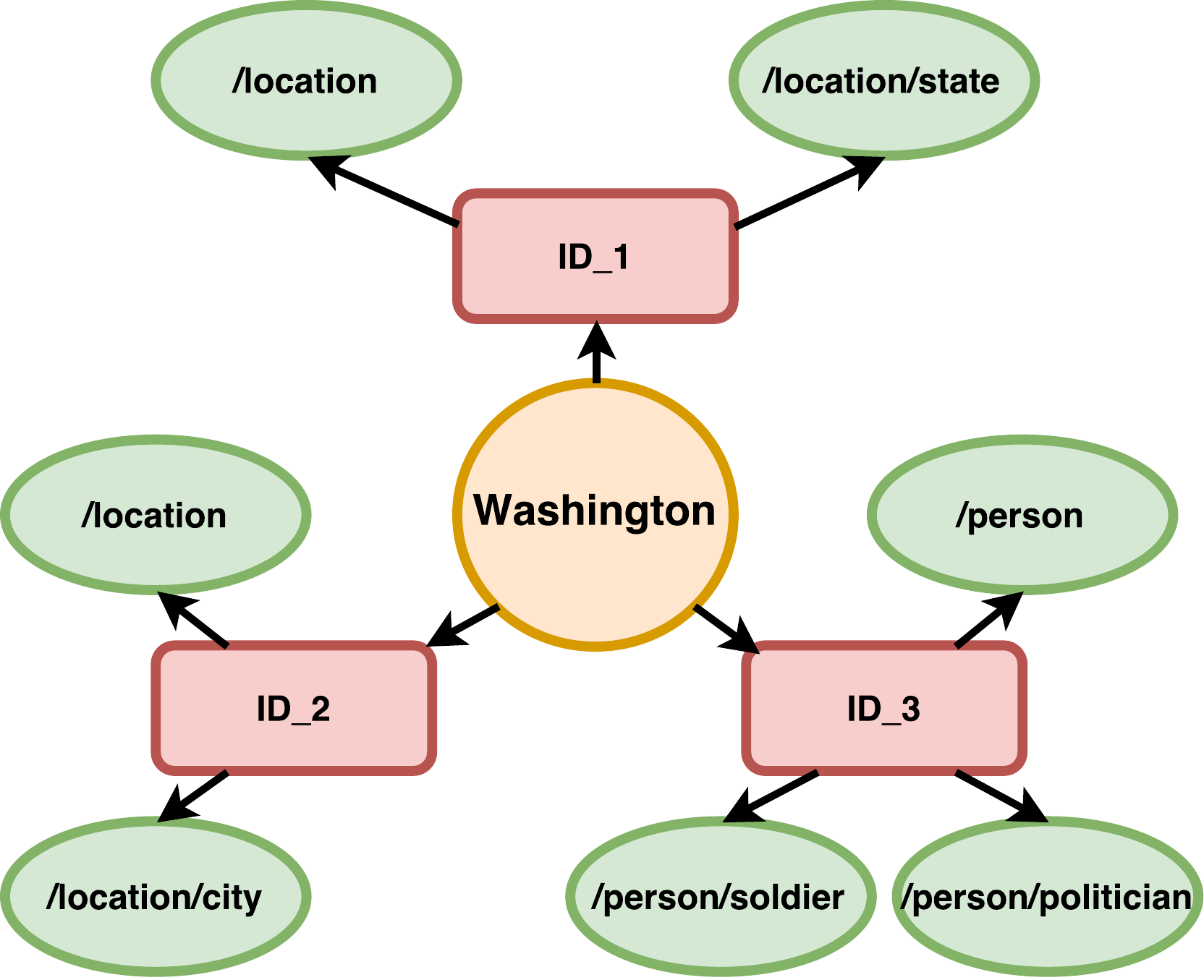}
\caption{Types (ellipses; green) of the entities (rectangles; red), to which the name ``Washington'' can refer.
Ideally, the embedding for ``Washington'' should represent all these types.}
\figlabel{n2i2t}
\end{figure}

Our evaluation is based on an entity-oriented task in information extraction (IE).
Different areas of IE try to predict relevant data about entities from text, either locally (i.e., at the context-level), or globally (i.e., at the corpus-level).
For example, local \cite{zeng14coling} and global \cite{Riedel13universal} in relation extraction, 
or local \cite{ling2012fine} and global \cite{figment15} in entity typing.
In most global tasks, each entity is indexed with an identifier (ID) that 
usually comes from knowledge bases such as Freebase. 
Exceptions are tasks in lexicon generation or population like entity set expansion (ESE) \cite{thelen2002bootstrapping}, which are global but without entity IDs. 
ESE usually starts from a few seed entities per set and completes the set 
using pattern-based methods.

Here, we address the task of \emph{fine-grained name typing \textbf{(FNT)}},
a global prediction task, operating on the surface names of entities. 
FNT and ESE share  applications in name lexicon population.
FNT is different from ESE because we assume to have sufficient training
instances for each type to train supervised models.

The challenging goal of FNT is to find the types of all entities a name can refer to. 
For example, "Washington" might refer to several entities which in turn may belong to multiple types, see \figref{n2i2t}.
In this  example, ``Washington''  refers to ``Washington DC (city)'',
``Washington (state)'', or ``George Washington (president)''.
Also, each entity can belong to several types, e.g., ``George Washington'' is a
\textsc{politician}, a \textsc{person} and a \textsc{soldier}, or
``Washington (state)'' is a \textsc{state} and a \textsc{location}.

Learning global representations for entities is very effective for
global prediction tasks in IE (cf., \newcite{figment15}).
For our task, FNT, we also learn a global representation for each name.
By doing so, we see this task as a challenging evaluation for embedding models.
We intend to use FNT to answer the following questions:
(i) How well can embeddings represent distinctive  information, i.e., different types or senses?
(ii) Which properties are important for an embedding model to do well on this task?

We build a novel large-scale dataset of (name, types) from Freebase with millions 
of examples.
The size of the dataset enables supervised approaches to work, an important requirement
to be able to look at different subspaces of embeddings
\cite{derata16}.
Also, in FNT names are---in contrast to concept categorization datasets---multi-labeled, which requires to look at multiple subspaces of embeddings.

In summary, our contributions are
(i) introducing a new evaluation method for word embeddings
(ii) publishing a new  dataset that is a good resource for evaluating word embeddings and is complementary to prior work:
it is very large, contains more different classes than previous word categorization datasets, and allows the direct evaluation of embeddings without confounding factors like sentence context\footnote{Our dataset is available at: \url{https://github.com/yyaghoobzadeh/name_typing}}. 

\section{Related Work}
\textbf{Embedding evaluation.}
\newcite{baroni14predict} evaluate embeddings on different intrinsic tests:
similarity, analogy, synonym detection, categorization and selectional preference.
\newcite{schnabel15evals} introduce 
tasks with more fine-grained datasets.
The concept categorization datasets used for embedding evaluation 
are mostly small ($<$500) \cite{baroni14predict}
and therefore measure the goodness of embeddings by the quality of their clustering.
In contrast, we test embeddings in a classification
setting and different subspaces of embeddings are analyzed.
Extrinsic evaluations 
are also used \cite{LiJ15a,koehn15,lai2015compoare}.
In most tasks, embeddings are used 
in context/sentence representations with composition involved. 
In this work, we evaluate embeddings in isolation, on their ability 
to represent multiple senses. 

\textbf{Related tasks and datasets.}
Our proposed task is fine-grained name typing (FNT).
A related task is entity set expansion (ESE): given a set of a few seed entities of a particular class, find other 
entities \cite{thelen2002bootstrapping,gupta14evalpatterns}.
We can formulate FNT as ESE, however, there is a difference in the training data assumption. 
For our task, we assume to have enough instances for each type available, and, therefore, to be able to use a supervised
learning approach. In contrast, for ESE, mostly only 3-5 seeds are given as
training seeds for a set, which makes an evaluation like ours impossible. 

\emph{Named entity recognition (NER)} consists of recognizing and classifying mentions of entities locally in a particular context
\cite{finkel2005}.
Recently, there has been increased interest in fine-grained typing of mentions \cite{ling2012fine,yogatama2015acl,xiang2016afet,attentiveTyper16}.
One way of solving our task is to collect every mention of a name, use NER to predict the context-dependent types of mentions,
and then take all predictions as the global types of the name.
However, our focus in this paper is on how embedding models perform
and propose this task as a good evaluation method.
We leave the comparison to an NER-based approach for future work.

\emph{Corpus-level fine-grained entity typing} is the task of predicting all types of \emph{entities} based on their mentions in a corpus \cite{figment15,figment17,figment18jair}.
This is similar to our task, FNT, but in FNT the goals is to find the corpus-level types of
\emph{names}.
Corpus-level entity typing has also been used for embedding evaluation \cite{derata16}.
However, they need an annotated corpus with entities. For FNT, however, pretrained word embeddings are sufficient for the evaluation.

Finally, there exists some previous work on FNT, e.g., 
\newcite{chesney2017}. 
In contrast to us, they do not explicitly focus on the evaluation of  embedding models, such that their 
dataset only contains a limited number of types. In contrast, we use 50 different types, making 
our dataset suitable for the type of evaluation intended.

\section{Multi-label Classification of Word
Embeddings}
Word embeddings are global representations of word properties  learned from the context 
distribution of words.
Words are usually ambiguous and belong to multiple
classes, e.g., multiple part-of-speech tags or multiple meanings. 
A good word  embedding should represent 
all information about the word, including its multiple classes.
  Our evaluation methodology is based on  this hypothesis and tries to test this through 
  multi-label classification of word embeddings. Here, we  focus on the semantic 
  property of nouns and entity names.
  We try to find all categories or types of a noun given its
embedding.

 Multi-label classification of embedding has
 multiple advantages over current evaluation methods: (i) large datasets can be created without
 much human annotation; (ii) more fine-grained
analysis of the results is possible through analyzing classification performance; (iii) it allows the direct evaluation of embeddings without confounding factors like sentence context.
  
\section{Fine-grained Name Typing}
We assume to have the following: a set of names $N$, a set of types $T$
and 
a membership function 
$m:  N \times T \mapsto \{0,1\}$ such that
$m(n,t)=1$ iff name $n$ has type $t$;
and a large corpus $C$.
In this problem setting, we address the task of
\emph{fine-grained name typing \textbf{(FNT)}}: we want to
infer from the corpus for each pair of name $n$ and 
type $t$ whether $m(n,t)=1$ holds.

For example, for the name 
``Hamilton'',  we should find all of the following: \textsc{location}, \textsc{organization}, \textsc{person}, \textsc{city}, \textsc{sports\_team} and \textsc{soldier}, since ``Hamilton'' 
can describe entities belonging to those types.
Another example is ``Falcon'' which is used for \textsc{animal}, \textsc{airplane}, \textsc{software}, \textsc{art}.
FNT sheds light on to which level these fine-grained types can be inferred from a corpus using embeddings.

\subsection{Embedding-based Model}
We aim to find $P(t|n)$, i.e., the probability of name $n$ having type $t$.
Given sufficient training instances for each type $t$, we can 
formulate the problem as a multi-label classification task.
As input, we use a representation for $n$, learned from the corpus $C$. 
Distributional representations have shown to capture various types of information about a word, especially their categories or types \cite{figment15}.

After learning an embedding for $n$, we train two kinds of binary classifiers for each type $t$ to to estimate $P(t|n)$:
 (i) linear: logistic regression (LR) with stochastic gradient decent; and
 (ii) non-linear: multi-layer perceptron (MLP) with one 
hidden layer and ReLU as the non-linearity.
We use the Scikit-learn \cite{sckit2011} toolkit for training our classifiers.

\section{Dataset}
Using Freebase \cite{bollacker2008freebase}, we first retrieve the set of all entities $E_n$ 
for each name $n$.\footnote{What we call ``names'' here are either \textit{names} or \textit{aliases} in the Freebase terminology.}
Then, we consider the types of all $e \in E_n$ the types of $n$.
See \figref{n2i2t} for an example: all of the shown types belong to the name "Washington".

Since some of the about 1,500 Freebase types have very few instances, we map them first to the
FIGER \cite{ling2012fine}  type-set, which contains 113  types. 
We then further restrict our set to the top 50 most frequent
types.
See \tabref{types} for the list of types.

\begin{table}
\small 
\begin{tabular}{p{3in}l}
/art, /art/film, /astral\_body, /biology, /broadcast\_network, /broadcast\_program, /building, /building/restaurant, /chemistry, /computer/programming\_language, /disease, /event, /food, /game, /geography/island, /geography/mountain, /god, /internet/website, /living\_thing, /location, /location/body\_of\_water, /location/cemetery, /location/city, /location/county, /medicine/drug, /medicine/medical\_treatment, /medicine/symptom, /music, /organization, /organization/airline, /organization/company, /organization/educational\_institution, /organization/sports\_team, /people/ethnicity, /person, /person/actor, /person/artist, /person/athlete, /person/author, /person/director, /person/engineer, /person/musician, /play, /product, /product/airplane, /product/instrument, /product/ship, /software, /title, /written\_work
\end{tabular}
\tablabel{types}
\caption{List of the 50 types in our FNT dataset.}
\end{table}

In order to be able to evaluate each embedding on its own, we
divide our dataset into single-word (891,241 names) and multi-word (8,907,715 names). 
In this work, the multi-word set is not used.
We then set a frequency threshold of 100 in our lowercased Wikipedia corpus \footnote{Our Wikipedia dump is from 2014.} and 
select randomly 100,000 of our dataset names that pass this threshold.
We then divide the names into train (50\%), dev (20\%) and test (30\%).
Some statistics of the single-word FNT dataset are shown in \tabref{dataset}.

\begin{table}[tb]
\footnotesize
\centering
\begin{tabular}{c|ccc} 
 & \#names         & avg \#types per name\\ 
\hline 
train & 50,000     & 3.78   \\
dev & 20,000       & 3.77  \\
test & 30,000      & 3.77  \\
\end{tabular} 
\caption{Some statistics (number of names; average number of types per name) 
for our name typing dataset.}
\tablabel{dataset}
\end{table}
\section{Experiments}
\label{sec:eval}
\subsection{FNT for Embedding Evaluation}

\textbf{Embedding models.} 
We choose four different embedding models for our comparisons: 
(i) SkipGram (henceforth SKIP) (skipgram bag-of-words model) \cite{mikolov2013efficient},
(ii) CBOW (continuous bag-of-words model) \cite{mikolov2013efficient},
(iii) Structured SkipGram (henceforth SSKIP)
\cite{ling15embeddings},
(iv) CWindow (henceforth CWIN) (continuous window model) \cite{ling15embeddings}.
SSKIP and CWIN are order-aware, i.e, they take the order of the context tokens
into account, while SKIP and CBOW are bag-of-words models.

\begin{table}[tb]
\footnotesize
\centering
\begin{tabular}{@{\hspace{0.0cm}}r@{\hspace{0.05cm}}l|cc|cc}
&& \multicolumn{2}{|c|}{LR} &
  \multicolumn{2}{|c}{MLP} \\
\hline 
&     & ACC  & Micro-F1  & ACC & Micro-F1 \\ 
\hline 
1&CBOW & 19.2  & 47.8 & 24.9 & \textbf{54.6} \\ 
2&SKIP & 22.6  & 49.3 & \textbf{25.2} & 53.5 \\ 
3&CWIN & 22.6 &  49.8 & 25.1 & 54.2 \\ 
4&SSKIP &\textbf{23.4} & \textbf{50.5}  & \textbf{25.2} & 53.6 \\ 
\end{tabular} 
\caption{Accuracy and micro-F1 results on FNT for different embedding models using two classifiers (LR and MLP).
Best result in each column is bold.
}
\tablabel{results}
\end{table}

\textbf{Results and analysis.}
We report the results for
 all embedding models using LR and MLP in \tabref{results}. We use the following evaluation measures,
 which are used in entity typing \cite{figment15}: (i) ACC (accuracy): percentage of test examples where all predictions are correct, (ii) Micro-F1: the global F1 computed over all the predictions.

 Models in lines 1-5 in Table 3 are trained on
 the Wikipedia corpus. We set the min frequency
 in corpus to 100. Window size = 3; negative sampling with $n=10$. Based on the results of LR, order-aware architectures are better than their bag-of-words
counterparts, i.e., SSKIP $>$ SKIP and CWIN $>$
 CBOW. Overall, SSKIP is the best using LR classification.
In MLP results, however, CBOW works best on micro-F1 measure and
SSKIP and SKIP are bests on accuracy.
There is no significant difference between CBOW and CWIN, 
or SSKIP and SKIP, respectively.
Overall, the nonlinear classifier (MLP) with one hidden layer outperforms the linear classifier (LR) substantially, emphasizing that the encoded information about different types is easier to extract with stronger models.

\textbf{Analysis on the number of name types. }
As a separate analysis, we measure how the classification performance depends on the $N$ number of types of a name.
To do so, we group test names based on their number of types. 
We keep the groups that have more than 100 members.
Then, we plot the F1 results of CBOW and CWIN models trained using MLP classifier in \figref{numtypes}.

As it is shown, both models get their best results on names with $N=2$.
We suppose that the bad performance of $N=1$ is related to the fact that one-type names have missing types in our dataset due to the incompleteness of Freebase.
The worse F1 of $N>=3$ compared to $N=2$ is expected since bigger $N$ means that the models need to predict more types from the name embeddings.
From $N=4$, somewhat surprisingly the F1 increases as $N$ increases.
This is perhaps related to the frequency of names
in the corpus, and its relation to the number of names types:
as $N$ increases, the frequency of words increases
and the embedding has a better quality.
However, this is only a hypothesis and more investigation is required. 
The other observation is in the trend of CBOW and CWIN results. CBOW is worse for $N<=2$, but it works clearly better for $N > 2$.
This shows that the embedding models behave differently for different number of classes they belong to.
This could also be related to the frequency of words.
Analysis of the reasons would be interesting. We leave it for the future work.

\begin{figure}
\includegraphics[width=200pt,height=125pt]{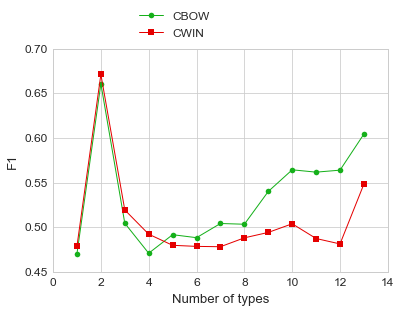}
\caption{Micro-F1 for names with different number of types.}
\figlabel{numtypes}
\end{figure}



\section{Conclusion}
We proposed multi-label classification of word  embeddings 
using the task of fine-grained 
typing of entity names. The dataset we  built is a resource 
that is complementary to prior  work in embedding evaluation: 
it is very large, its  examples are multi-labeled with very fine-grained 
 classes, and it allows the direct evaluation of embeddings without 
the need for context. 
We analyzed the performance of different embedding 
 models on this dataset, showing differences in  their performance as well 
as some of their limits  in representing types accurately and completely.

More analysis and evaluation is necessary though, but we believe by using this 
kind  of dataset, we are able to do much more than what we could do before with the small manually built word similarity and categorization benchmarks.

\newcommand\BibTeX{B{\scib}\TeX}

\bibliography{ref}
\bibliographystyle{acl_natbib}

\end{document}